# Bayesian Hierarchical Mixtures of Experts

**Christopher M. Bishop    Markus Svensén**
Microsoft Research
7 J J Thomson Avenue
Cambridge, CB3 0FB, U.K.
http://research.microsoft.com/~cmbishop/

## Abstract

The Hierarchical Mixture of Experts (HME) is a well-known tree-structured model for regression and classification, based on soft probabilistic splits of the input space. In its original formulation its parameters are determined by maximum likelihood, which is prone to severe over-fitting, including singularities in the likelihood function. Furthermore the maximum likelihood framework offers no natural metric for optimizing the complexity and structure of the tree. Previous attempts to provide a Bayesian treatment of the HME model have relied either on local Gaussian representations based on the Laplace approximation, or have modified the model so that it represents the joint distribution of both input and output variables, which can be wasteful of resources if the goal is prediction. In this paper we describe a fully Bayesian treatment of the original HME model based on variational inference. By combining 'local' and 'global' variational methods we obtain a rigorous lower bound on the marginal probability of the data under the model. This bound is optimized during the training phase, and its resulting value can be used for model order selection. We present results using this approach for data sets describing robot arm kinematics.

## 1 INTRODUCTION

The *hierarchical mixture of experts* (HME) is a parametric probabilistic model for solving regression and classification problems (Jordan and Jacobs 1994). The HME can be viewed as a conditional mixture model in which the distribution of the target variables is given by a mixture of component distributions in which the components, as well as the mixing coefficients, are conditioned on the input variables. The component distributions are referred to as *experts*, while mixing coefficients are controlled by *gating distributions*. Values for the model parameters can be set using maximum likelihood, for which there exists an efficient EM algorithm (Jordan and Jacobs 1994). Such a model will automatically perform a *soft* partitioning of the data set into groups corresponding to different regions of input space and simultaneously fit separate models (corresponding to the mixture components) to each of those groups.

A major limitation of the maximum likelihood approach is the propensity for over-fitting. This can be particularly problematic in a complex model such as the HME due to the relatively large number of parameters involved in defining the expert and gating distributions. Indeed, there are many singularities in the likelihood function arising whenever one of the mixture components 'collapses' onto a single data point. Furthermore, the maximum likelihood framework provides no direct mechanism for determining either the number of nodes in the HME tree, or its topology, since optimization of the likelihood function will simply favour ever more complex models. Both of these problems can be resolved by adopting a Bayesian approach, in which we introduce prior distributions over the parameters of the HME. However, an exact Bayesian treatment of the HME is intractable. In fact the gating distributions do not even admit conjugate priors.

Currently there is considerable interest in deterministic approximation schemes for Bayesian inference based on variational methods. An application of variational inference to the HME model was previously investigated by Waterhouse, MacKay, and Robinson (1996). However, in order to define a tractable algorithm they fitted Gaussian distributions over the parameters controlling the gating functions using the Laplace approximation. This sacrifices one of the most appealing aspects of the variational approach namely that it optimizes a rigorous lower bound on the log marginal likelihood.

A variational treatment for a related model was recently given by Ueda and Ghahramani (2002), in which they obtain tractability by considering a model which represents the *joint* distribution over both input and output variables (though in fact they only implement a single-layer model, not a hierarchical version). If the goal is prediction then such an approach can be very wasteful of resources, as well as demanding of data, since the distribution over the input space (which often has much higher dimensionality than the target space) is not required.

Here we build on recent developments in variational methods to provide a fully Bayesian treatment of the hierarchical mixture of experts model in which we optimize a well defined lower bound on the log marginal probability of the observed data (Bishop 2002).

We illustrate this framework by applying the Bayesian HME to example problems involving the kinematics of robot arms. For the case of inverse kinematics the conditional distribution being modelled is multi-modal, and this is handled well by the HME approach.

## 2 THE HME MODEL

The HME describes a conditional probability distribution over a vector t of target variables, conditioned on a vector x of inputs. For a given value of x, the distribution over t is a mixture distribution in which the mixing coefficients are defined with the help of a tree-structured graph, of which a simple example is shown in Figure 1. Each expert rep-resents a probability distribution over t, conditioned on the input vector x. The gating nodes are probabilistic switches which decide which of the expert nodes is selected, and again these switching probabilities are functions of x. For the moment let us suppose t has real valued components and that the corresponding expert distributions are Gaussian. Relaxations of the Gaussian assumption will be discussed later. We shall also suppose that the HME tree is binary.

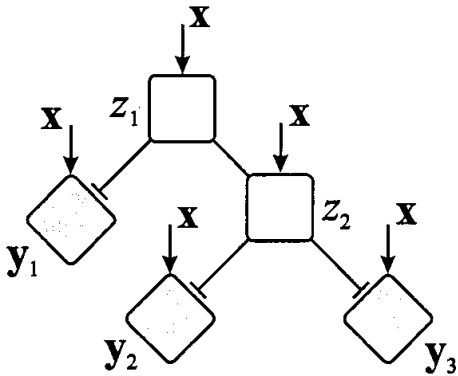

Figure 1: A hierarchical mixture of experts, comprising *expert nodes*, shown as diamonds, and *gating nodes*, shown as squares. The $z_i$ denote the binary variables associated with the gating nodes whereas the $y_j$ denote the means of the conditional distributions over the target variable t.

The conditional distribution for expert $j$ is a Gaussian with mean $y_j(x) = W_j x$, so that

$$p(t|x, W_j, \tau_j) = \mathcal{N}(t|W_j x, \tau_j^{-1} I) \qquad (1)$$

where $\mathcal{N}(t|\mu, \Sigma)$ denotes a Gaussian distribution with mean $\mu$ and covariance $\Sigma$. Here $W_j$ is a matrix of parameters associated with expert $j$, $\tau_j$ is the precision (inverse variance) of the distribution, and $I$ is the unit matrix. In order to simplify the notation we have assumed that the vector of inputs has been augmented with an additional dummy input variable whose value is clamped to 1, so that the corresponding column of $W$ represents a 'bias'.

The HME model is perhaps best understood generatively. Each gating node has an associated binary variable $z_i \in \{0, 1\}$, whose value is chosen with probability given by

$$p(z_i|x, v_i) = \sigma(v_i^T x)^{z_i} [1 - \sigma(v_i^T x)]^{1-z_i} \qquad (2)$$

where

$$\sigma(a) = \frac{1}{1 + \exp(-a)}$$

is the logistic sigmoid function, and $v_i$ is a vector of parameters governing the distribution. If $z_i = 1$ we go down the left branch while if $z_i = 0$ we go down the right branch. Starting at the top of the tree, we thereby stochastically choose a path down to a single expert node $j$, and then generate a value for t from conditional distribution for that expert.

We see that, given the states of the gating variables, the HME model corresponds to a conditional distribution for t of the form

$$p(t|x, W, \tau, z) = \prod_{j=1}^{M} \mathcal{N}(t|W_j x, \tau_j^{-1} I)^{\zeta_j}$$

where $M$ is the total number of experts, $W$ denotes $\{W_j\}$, and $\tau$ denotes $\{\tau_j\}$. Here we have defined

$$\zeta_j = \prod_i \tilde{z}_i, \qquad (3)$$

in which the product is taken over all gating nodes on the unique path from the root node to the $j$th expert, and

$$\tilde{z}_i = \begin{cases} z_i & \text{if } j \text{ is in the left sub-tree of } i, \\ 1 - z_i & \text{otherwise.} \end{cases}$$

Marginalizing over the gating variables $z = \{z_i\}$ we obtain

$$\begin{aligned} &p(t|x, W, v, \tau) \\ &= \sum_z \prod_j \mathcal{N}(t|W_j x, \tau_j^{-1} I)^{\zeta_j} \prod_i p(z_i|x, v_i) \\ &= \sum_j \pi_j(x) \mathcal{N}(t|W_j x, \tau_j^{-1} I) \end{aligned}$$

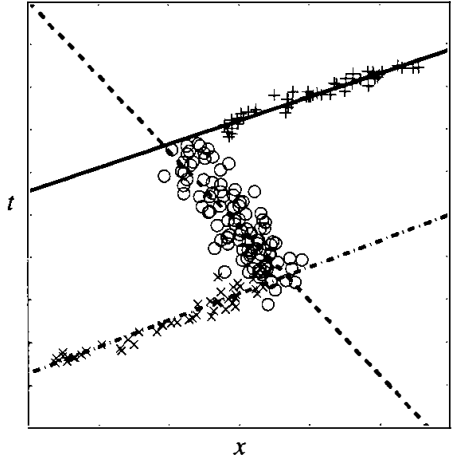

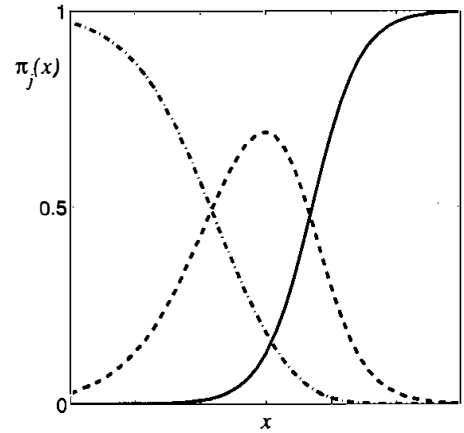

Figure 2: A toy data set together with the result of fitting the HME model of Figure 1 using the Bayesian technique described in this paper. The data points have been plotted with symbols indicating which expert is most likely to have generated the points; + indicates the expert drawn with a solid line ($y_1$ in Figure 1), × indicates the expert drawn with a dash-dotted line ($y_2$ in Figure 1) and ○ indicates the expert drawn with a dashed line ($y_3$ in Figure 1).

Figure 3: Mixing coefficients for each of the three experts as a function of $x$, for the toy problem shown in Figure 2.

We illustrate the HME model by considering a simple toy problem (Bishop 1995) shown in Figure 2. Here we have generated 200 values of $t$ uniformly from the interval $(0, 1)$ and then evaluated $x = t + 0.3\sin(2\pi t) + \epsilon$ where $\epsilon$ is a zero-mean Gaussian random variable having standard deviation 0.05. Then we learn the *inverse* of this problem, namely that of predicting $t$ given a new value of $x$, using an HME model having 2 gating nodes and 3 experts, with the architecture shown in Figure 1. The lines in Figure 2 correspond to the mean outputs of the three experts. In Figure 3 we show the (means of the distribution over the) mixing coefficients as a function of $x$ for the three experts.

so that the conditional distribution $p(t|x, \mathbf{W}, \mathbf{v}, \tau)$ is a mixture of Gaussians in which the mixing coefficient $\pi_j(\mathbf{x})$ for expert $j$ is given by a product over all gating nodes on the unique path from the root to expert $j$ of factors $\sigma(\mathbf{v}_i^T \mathbf{x})$ or $1 - \sigma(\mathbf{v}_i^T \mathbf{x})$ according to whether the branch at the $i$th node corresponds to $z_i = 1$ or $z_i = 0$.

A key feature to note is that this conditional distribution is multi-modal. This is possible because the gating node outputs are smooth functions of the input variable. Such multi-modality, which often arises in the solution of *inverse problems*, could *not* be captured in CART (Classification and Regression Trees) or similar models, since they assign each point of the input space to one, and only one, of the terminal nodes ('hard' splits).

If we are given an i.i.d. data set $\mathbf{X} = \{\mathbf{x}_n\}$, comprising $N$ observations of the input vector $\mathbf{x}$, and corresponding observations of the target vector $\mathbf{T} = \{\mathbf{t}_n\}$, the likelihood function is given by

$$L(\mathbf{W}, \mathbf{v}, \tau) = \prod_{n=1}^{N} p(\mathbf{t}_n | \mathbf{x}_n, \mathbf{W}, \mathbf{v}, \tau).$$

### 2.1 A Bayesian HME

Note that there will be a separate latent variable $\mathbf{z}_n$ for each data point, and in the likelihood function we are implicitly marginalizing over $\mathbf{Z} = \{\mathbf{z}_n\}$. In its original formulation, the parameters $\mathbf{W}$, $\tau$ and $\mathbf{v}$ of the HME were determined by maximum likelihood using the EM algorithm, in which the E-step involves finding the posterior distribution over the $\{\mathbf{z}_n\}$, and the M-step involves maximizing the corresponding expected complete-data log likelihood with respect to $\mathbf{W}$, $\mathbf{v}$ and $\tau$.

We can avoid the severe bias of maximum likelihood, and also obtain a principled framework for optimizing the complexity and topology of the HME graph, by adopting a Bayesian treatment. Specifically we define a Gaussian prior distribution independently over of the parameters $\mathbf{v}_i$ for each of the gating nodes given by

$$p(\mathbf{v}_i | \beta_i) = \mathcal{N}(\mathbf{v}_i | \mathbf{0}, \beta_i^{-1} \mathbf{I}).$$

Similarly, for the parameters $\mathbf{W}_j$ of the expert nodes we define priors given by

$$p(\mathbf{W}_j | \alpha_j) = \prod_{k=1}^{d} \mathcal{N}(\mathbf{w}_{jk} | \mathbf{0}, \alpha_j^{-1} \mathbf{I})$$

Note that the maximum likelihood solution for $\mathbf{W}_j$ of a conditional Gaussian distribution of the form (1) corresponds to linear regression, so that the HME model is performing a soft partitioning of the data set and then fitting a linear regression model to each of those partitions separately.

where $k$ runs over the target variables, and $d$ is the dimensionality of the target space. Thus the rows $\mathbf{w}_{jk}$ of $\mathbf{W}_j$, corresponding to different target variables, are given independent priors. The hyper-parameters $\beta_i$ and $\alpha_j$, as well as the noise precisions $\tau_j$, are given conjugate gamma distributions

$$p(\alpha_j) = \text{Gam}(\alpha_j|a,b)$$
$$p(\beta_i) = \text{Gam}(\beta_i|a,b)$$
$$p(\tau_j) = \text{Gam}(\tau_j|a,b)$$

where

$$\text{Gam}(\tau|a,b) \equiv \frac{b^a \tau^{(a-1)} \exp(-a\tau)}{\Gamma(a)}.$$

in which we set $a = 10^{-2}$ and $b = 10^{-4}$ giving broad hyper-priors. Our Bayesian HME can be expressed as the directed probabilistic graphical model shown in Figure 4. Exact inference in this model is not analytically

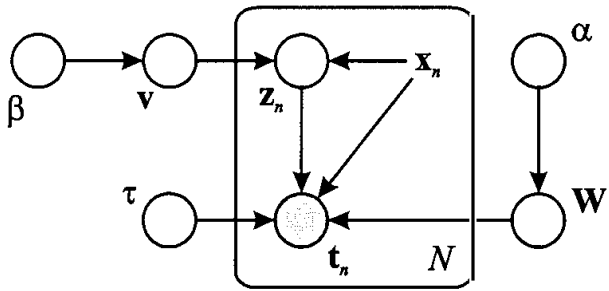

Figure 4: Graphical model representation of the Bayesian HME. The box, called a *plate*, denotes $N$ copies of the nodes shown inside the box. This model does not include the distribution over the input variables $\mathbf{x}_n$ and so these do not have a corresponding stochastic node. The output node $\mathbf{t}_n$ is shaded, indicating that these variables are observed.

tractable, and so we make use of variational methods (Jordan, Ghahramani, Jaakkola, and Saul 1998).

## 3 VARIATIONAL INFERENCE

Our goal is to find a variational distribution $q(\mathbf{U})$ that approximates the true posterior distribution $p(\mathbf{U}|\mathbf{T})$, where we collectively denote the hidden variables by $\mathbf{U} = (\mathbf{W}, \boldsymbol{\tau}, \mathbf{Z}, \mathbf{v}, \boldsymbol{\alpha}, \boldsymbol{\beta})$, and we suppress the dependence on $\mathbf{X}$. To do this we note the following decomposition of the log marginal probability of the observed data, which holds for any choice of distribution $q(\mathbf{U})$

$$\ln p(\mathbf{T}) = \mathcal{L}(q) + \text{KL}(q\|p) \qquad (4)$$

where

$$\mathcal{L}(q) = \int q(\mathbf{U}) \ln\left\{\frac{p(\mathbf{U},\mathbf{T})}{q(\mathbf{U})}\right\} d\mathbf{U} \qquad (5)$$

$$\text{KL}(q\|p) = -\int q(\mathbf{U}) \ln\left\{\frac{p(\mathbf{U}|\mathbf{T})}{q(\mathbf{U})}\right\} d\mathbf{U} \qquad (6)$$

and the integrals are replaced by sums in the case of the discrete variables in $\mathbf{Z}$. Here $\text{KL}(q\|p)$ is the Kullback-Leibler divergence between the variational distribution $q(\mathbf{U})$ and the true posterior $p(\mathbf{U}|\mathbf{T})$. Since this satisfies $\text{KL}(q\|p) \geqslant 0$ it follows from (4) that the quantity $\mathcal{L}(q)$ forms a lower bound on $\ln p(\mathbf{T})$. Maximizing the lower bound with respect to $q(\mathbf{U})$ is equivalent to minimizing the Kullback-Leibler divergence, which has the effect of bringing the variational distribution closer to the true posterior.

### 3.1 Local Convex Bound

Our goal is to find a variational distribution $q(\mathbf{U})$ which will give a tight lower bound, yet which is sufficiently simple that it remains tractable. Our approach will be based on factorized forms for the variational distribution, as discussed in Section 3.2. While this approach is widely used (Bishop, Spiegelhalter, and Winn 2002), and has given good results for a wide range of models, it does not directly lead to a tractable solution for the Bayesian HME.

The difficulty lies with the sigmoid function in (2) which spoils the conjugate-exponential structure of the model. In this paper we address this problem using another technique from the field of variational methods based on bounding log convex functions (Jaakkola and Jordan 2000).

We first of all re-write (2) in the form

$$p(z_i|\mathbf{v}_i, \mathbf{x}) = \exp(z_i \mathbf{v}_i^T \mathbf{x}) \sigma(-\mathbf{v}_i^T \mathbf{x}).$$

Next we make use of a variational bound for the logistic sigmoid function in the form

$$\sigma(x) \geqslant F(x,\xi) \equiv \sigma(\xi) \exp\left\{(x-\xi)/2 - \lambda(\xi)(x^2 - \xi^2)\right\} \qquad (7)$$

where $\lambda(\xi) = \tanh(\xi/2)/(4\xi)$, and $\xi$ is a variational parameter. For any given value of $x$ we can make this bound exact by an appropriate choice of the variational parameter $\xi$, namely $\xi = x$. In fact the bound is exact at both $x = \xi$ and $x = -\xi$. The bound is illustrated in Figure 5, in which the solid curve shows the logistic sigmoid function $\sigma(x)$, and the dashed curve shows the lower bound $F(x,\xi)$.

We can use this result to derive a new bound $\widetilde{\mathcal{L}} \leqslant \mathcal{L}$ which is obtained by replacing every occurrence of $p(z_i|\mathbf{v}_i, \mathbf{x})$ with its lower bound $\exp(z_i \mathbf{v}_i^T \mathbf{x}) F(-\mathbf{v}_i^T \mathbf{x}, \xi_i)$, where $F(\cdot, \cdot)$ is defined by (7). So far as the dependence on $\mathbf{v}$ is concerned, the effect of this transformation is to replace the logistic sigmoid with an exponential, thereby restoring conjugacy to the Bayesian model. For each gating node $i$ there is a separate variational parameter $\xi_{in}$ for each observation $n$, and the values of these parameters can be optimized to yield the tightest bound.

Note that the variational bound given by $F(x,\xi)$ does not extend to multi-way gating nodes governed by softmax functions. However, a complex, multi-way division of the input space can be represented using binary splits provided the tree structure is sufficiently rich.

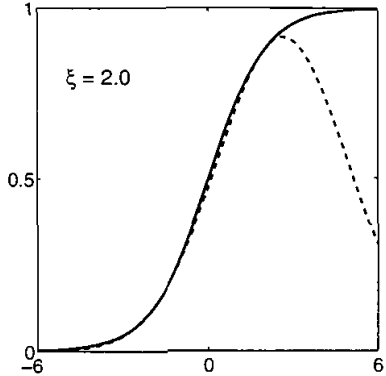

Figure 5: Logistic sigmoid function and variational bound.

### 3.2 Factorized Distributions

We now choose some family of distributions to represent $q(\mathbf{U})$ and then seek a member of that family which maximizes the lower bound $\mathcal{L}(q)$. If we allow $q(\mathbf{U})$ to have complete flexibility then we see that the maximum of the lower bound occurs for $q(\mathbf{U}) = p(\mathbf{U}|\mathbf{T})$ so that the variational posterior distribution equals the true posterior. In this case the Kullback-Leibler divergence vanishes and $\mathcal{L}(q) = \ln p(\mathbf{T})$. However, working with the true posterior distribution is computationally intractable. We must therefore consider a more restricted family of $q$ distributions which has the property that the lower bound (5) can be evaluated and optimized efficiently and yet which is still sufficiently flexible as to give a good approximation to the true posterior distribution.

Here we consider the set of distributions which factorizes with respect to disjoint groups $\mathbf{U}_i$ of variables

$$q(\mathbf{U}) = \prod_i q_i(\mathbf{U}_i). \quad (8)$$

Substituting (8) into (5) we can maximize $\mathcal{L}(q)$ variationally with respect to one of the factors, say $q_i(\mathbf{U}_i)$ keeping all $q_j$ for $j \neq i$ fixed. This leads to the solution

$$\ln q_i^*(\mathbf{U}_i) = \langle \ln p(\mathbf{U}, \mathbf{T}) \rangle_{\{j \neq i\}} + \text{const}. \quad (9)$$

where $\langle \cdot \rangle_k$ denotes an expectation with respect to the distribution $q_k(\mathbf{U}_k)$, and the constant represents the log of the normalization coefficient for the distribution. In the case of models having a conjugate-exponential structure, we can evaluate the right hand side explicitly and obtain a solution for $q_i^*(\mathbf{U}_i)$ which belongs to the same class of distribution (for instance Gaussian or Gamma) as the original conditional $p(\mathbf{U}_i|\cdot)$.

Note that these are coupled equations since the solution for each $q_i(\mathbf{U}_i)$ depends on expectations with respect to the other factors $\{q_{j \neq i}\}$. The variational optimization proceeds by initializing the $q_i(\mathbf{U}_i)$ and then cycling through each factor in turn replacing the current distribution with a revised estimate given by (9).

For the Bayesian HME model we consider the specific factorization given by

$$q(\mathbf{U}) = q_{\mathbf{W}}(\mathbf{W})q_{\mathcal{T}}(\tau)q_{\mathbf{Z}}(\mathbf{Z})q_{\mathbf{v}}(\mathbf{v})q_{\alpha}(\alpha)q_{\beta}(\beta), \quad (10)$$

from which we obtain a set of re-estimation equations for each of the factors. For instance, the optimal solution for $q_{\mathbf{Z}}(\mathbf{Z})$ takes the form

$$q_{\mathbf{Z}}^*(\mathbf{Z}) = \prod_n \prod_i \sigma(h_{in})^{z_{in}}[1 - \sigma(h_{in})]^{1-z_{in}}$$

where the product over $i$ runs over all gating nodes, and

$$h_{in} = \sum_j \zeta_{jn}^{\not i} \left( \frac{D}{2}\langle \ln \tau_j \rangle - \frac{\langle \tau_j \rangle}{2}\langle \|\mathbf{t}_n - \mathbf{W}_j\mathbf{x}_n\|^2 \rangle \right) + \langle \mathbf{v}_i^{\mathrm{T}} \rangle \mathbf{x}_n$$

in which $D$ is the dimensionality of the target space, and $\zeta_{jn}^{\not i}$ has an analogous definition to $\zeta_j$ in (3) but with the $i$th term omitted. Note that the solution for $q_{\mathbf{Z}}^*(\mathbf{Z})$ depends on moments, such as $\langle \mathbf{v}_i^{\mathrm{T}} \rangle$, evaluated with respect to other factors in the variational $q$ distribution. Similar results are obtained for the other factors, in which the solutions for $q_\alpha$, $q_\beta$ and $q_\tau$ are gamma distributions while those for $q_{\mathbf{W}}$ and $q_{\mathbf{v}}$ are Gaussian. Due to lack of space we do not reproduce all of the update equations here.

Optimization of the $\xi$ parameters is achieved by maximizing the lower bound on the marginal likelihood, leading to the re-estimation equations

$$\xi_{in}^2 = \mathbf{x}_n^{\mathrm{T}} \langle \mathbf{v}_i \mathbf{v}_i^{\mathrm{T}} \rangle \mathbf{x}_n.$$

Re-estimation of the $\{\xi_{in}\}$ is interleaved with re-estimation of the factors in the variational posterior.

It should be noted that, although we are optimizing a well defined bound on the log marginal likelihood, we will converge to a local, but not necessarily a global, maximum. We address this through multiple re-starts with random initialization of the variational distribution.

### 3.3 Lower Bound

In this variational framework it is also tractable to compute the value of the lower bound $\widetilde{\mathcal{L}}$ itself (Bishop, Spiegelhalter, and Winn 2002). We omit detailed expressions due to lack of space. In fact most of the terms which appear in the bound are already evaluated during the variational updates, so little additional computational cost is incurred. Evaluation of the bound can be used to monitor convergence and to set stopping criteria.

The bound also provides a check on the correctness of the algorithm and its implementation since each variational update should not lead to a decrease in the value of this bound. As a further check on the correctness of the implementation during the debugging phase, we use finite differences to evaluate the derivatives with respect to each set of parameters immediately after updating the corresponding factor in the variational posterior distribution, to confirm that a local maximum with respect to those parameters has indeed been reached.

If we consider a range of models indexed by $\mathcal{M}$ then the posterior distribution over models, given an observed data set $\mathbf{T}$, is given by $p(\mathcal{M}|\mathbf{T}) \propto p(\mathcal{M})p(\mathbf{T}|\mathcal{M})$ where $p(\mathcal{M})$ is a prior distribution over models. This posterior distribution can be used to select the most probable model, or to perform model averaging. In contrast to the maximum likelihood approach, which always favours ever more complex models, the Bayesian posterior provides a natural trade-off between fitting the data and model complexity.

The key quantity we need to evaluate is therefore the model 'evidence' $p(\mathbf{T}|\mathcal{M})$, whose logarithm we have approximated through the lower bound, $\mathcal{L}_\mathcal{M}$, in the form (5). In order to make effective use of the bound, however, it is important to obtain good solutions to the variational equations by avoiding poor local maxima.

For moderately sized trees, we can determine the architecture of the HME by simply evaluating exhaustively all possible trees up to some maximum depth, that are unique up to symmetry. For each architecture we perform multiple training runs using different random initializations and keep only the one for which the resulting value of the lower bound is largest, since this represents our best approximation to the posterior distribution. These largest values are then compared and the largest of these is used to determine the choice of architecture. Note that the largest values from each model could also be used to construct (unnormalized) weights $\exp(\mathcal{L}_\mathcal{M})$ for use in model averaging. For larger trees, we could consider using greedy search algorithms (Ueda and Ghahramani 2002) or Monte-Carlo methods (Chipman, George, and McCulloch 2002).

The application of the lower bound in model order selection can be illustrated using the toy data set of Figure 2. Here we consider HME models having between 2 and 5 expert nodes, and for each architecture we perform 100 runs of the variational optimization starting with random initializations. Plots of the resulting values of the lower bound are shown in Figure 6. We see that there are many local maxima of the lower bound. Also we observe that the largest value of the lower bound for each architecture exhibits the classical 'Ockham hill', which has its maxi-mum value for the optimal architecture (3 experts in this case) and falls steeply for models which are less complex (in this case ones having 2 experts) due to the poor fit to the data, and also falls away, but much less sharply, for more complex models as any improvement in data fit is offset by an increasing complexity penalty arising from the Bayesian marginalization.

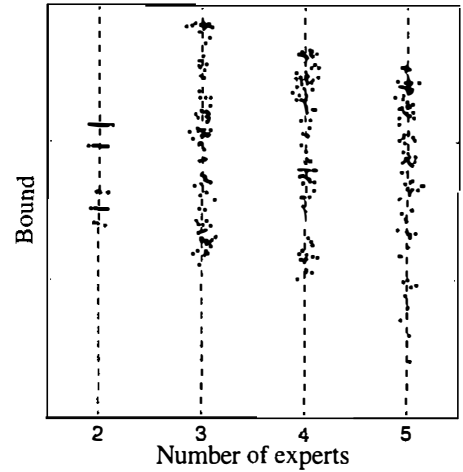

Figure 6: Plot of the the lower bound values obtained with multiple random starts for HME architectures having various numbers of expert nodes, applied to the toy data set of Figure 2.

## 4 RESULTS

We illustrate the application of the Bayesian approach to the HME using a data set derived from the kinematics of a two-link robot arm, whose geometry is shown in Figure 7. The cartesian coordinates of the robot end effector

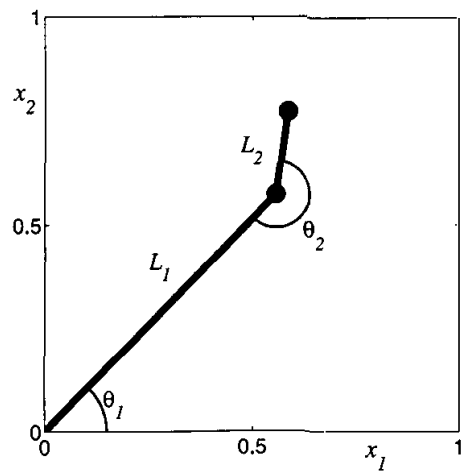

Figure 7: Geometry of the two-link robot arm used to generate data to illustrate the Bayesian HME model.

are given by the *forward kinematics* equations

$$x_1 = L_1 \cos\theta_1 - L_2 \cos(\theta_1 + \theta_2)$$
$$x_2 = L_1 \sin\theta_1 - L_2 \sin(\theta_1 + \theta_2)$$

where $L_1$ and $L_2$ are the lengths of the links, and $\theta_1$ and $\theta_2$ are the joint angles. Note that the forward kinematic equations have a unique solution for given values of the joint angles. However, we are interested in solving the *inverse kinematics* in which we are given the end effector location and have to determine the corresponding joint angles. This inverse problem can be multimodal due to the presence of two solutions, as indicated in Figure 8.

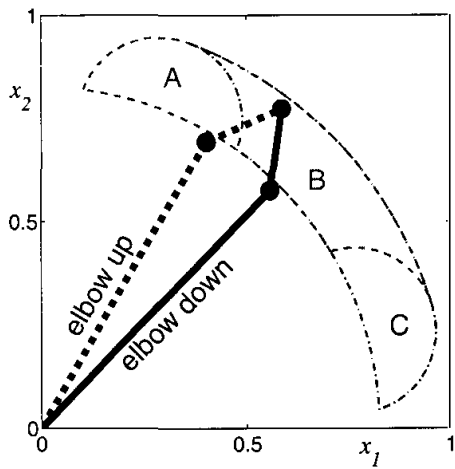

Figure 8: Illustration of the regions of space covered by the robot end effector for given ranges of the joint angles. Regions A and B are accessible in an elbow down configuration while regions B and C are accessible in elbow up configurations. Thus in region B there are two possible solutions to the inverse kinematics problem.

Here we consider joint lengths $L_1 = 0.8$ and $L_2 = 0.2$, and we limit the joint angles to the ranges $0.3 \leqslant \theta_1 \leqslant 1.2$ and $\pi/2 \leqslant \theta_2 \leqslant 3\pi/2$. This allows the robot end effector to sweep out the regions shown in Figure 8.

Standard approaches to regression, based on least squares optimization, can give extremely poor results when applied to multi-modal problems. Figure 9 shows the result of training a multi-layer perceptron neural network on this data set using least squares. The network had 20 hidden units and was trained using 3000 iterations of conjugate gradients. We see that the results are particularly poor in the central region where the inverse kinematics is bimodal. This arises because a least squares solution is computing a conditional average, and the average of the two solutions is not itself a solution (in fact it corresponds to a solution with $\theta_2 = \pi$ in which the robot arm is 'straight', hence the appearance of radial lines in the central region in Figure 9).

The corresponding results obtained with the Bayesian

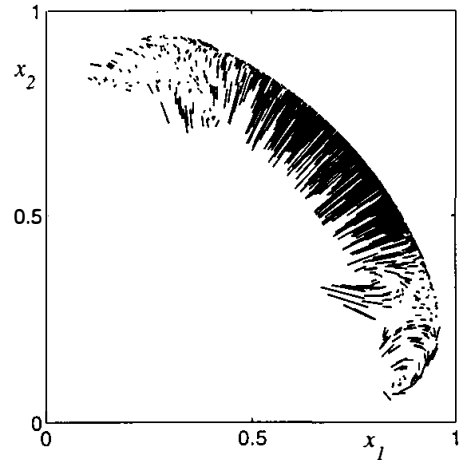

Figure 9: Test set results for the two-link robot arm problem obtained using a neural network trained by least squares. For each test set point we have drawn a line between the end effector position (which is the input to the neural network) and the corresponding predicted position obtained by taking the joint angles output by the network and feeding them through the forward kinematic equations. Thus the length of a line indicates the magnitude of the corresponding predictive error.

HME model are shown in Figure 10. Here an HME model with 16 experts was trained with 100 random starts and the solution giving the largest value for the lower bound was chosen and used to generate the plot. For each test input, the prediction is given by the mean of the expert distribution for the most probable expert. Thus in the multi-modal region the model selects either one branch or the other, not their average, and hence the predictive errors are much smaller.

As a third application of the Bayesian HME model we consider a more realistic robot arm problem, taken from a family of public domain data sets which have been synthetically generated to model the forward kinematics of an 8-link all-revolute robot arm. The task is to predict the distance between the end effector of the robot arm and a specified point, from the parameters and angles of the robot arm. This family offers data sets with varying number of input parameters, degree of non-linearity and level of noise. It is available, together with more detailed documentation, from the Delve repository[1].

We used the data set with 8 inputs, a high degree of non-linearity and medium noise (kin-8nm). The size of the training set was 1024 and the data was normalized to zero mean and unit variance. We trained HME models with 2–8 experts, trying all unique tree topologies and for each instance trying 50 different random starts. In an attempt to avoid local optima, we used a deterministic annealing schedule during the training, scaling the conditional probability of the data with an inverse temperature parameter. This parameter had an initial value of 5.85, which reduced with a factor of 0.97 at each of the 200 first iterations of training, which were followed by another 600 iterations of training with the temperature fixed at 1.0.

---

[1] http://www.cs.toronto.edu/~delve/

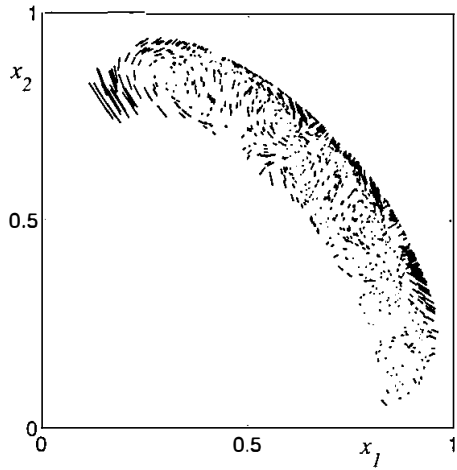

Figure 10: Test set results, analogous to those in Figure 9, but obtained using the Bayesian HME. Note the much smaller errors particularly in the central multi-modal region.

We selected the model with the overall highest bound, which turned out to be a model with four experts, and then evaluated this model using an independent test set from the same problem of size 1024. The standardized mean squared error (MSE) on this test set was 0.249, which compares well with the results presented by Waterhouse (1997) for a range of different HME models, which had MSE values of 0.262–0.378[2].

## 5 DISCUSSION

In this paper we have presented a variational treatment for a Bayesian hierarchical mixture of experts model which maintains a rigorous lower bound on the log marginal likelihood. We have shown that the model can learn good solutions to multi-dimensional regression problems, and that the lower bound can be used to perform model selection.

Although we have focussed on regression problems in this paper, it is straightforward to apply this approach to binary classification problems for a model with logistic sigmoid experts simply by applying the variational bound (7) to the expert nodes as well as to the gating nodes.

---

[2] Of all models evaluated by Waterhouse (1997), the lowest MSE, 0.094, was obtained for an MLP, whereas a linear regression model scored worst with a MSE of 0.569.

In common with other applications of variational inference we have observed that the lower bound possesses many local maxima, not all of which represent good solutions. Here we have proceeded by using multiple runs with random initializations and selecting the best optima, augmented where necessary by deterministic annealing. For large data sets and complex models this may be computationally infeasible, and it remains an open research issue to find effective and broadly applicable methods to address the local maxima problem for variational methods.

### Acknowledgements

We would like to thank Zoubin Ghahramani, Tommi Jaakkola and Michael Jordan for valuable discussions relating to this work.

### References

Bishop, C. M. (1995). *Neural Networks for Pattern Recognition*. Oxford University Press.

Bishop, C. M. (2002). Discussion of Chipman, George, and McCulloch (2002), pp. 99–103,

Bishop, C. M., D. Spiegelhalter, and J. Winn (2002). VIBES: A variational inference engine for Bayesian networks. In *Advances in Neural Information Processing Systems*, Volume 15. To appear.

Chipman, H. A., E. I. George, and R. E. McCulloch (2002). Bayesian treed generalized linear models. In J. M. Bernardo (Ed.), *Proceedings Seventh Valencia International Meeting on Bayesian Statistics*, pp. 85–98. Oxford University Press. To appear.

Jaakkola, T. and M. I. Jordan (2000). Bayesian parameter estimation through variational methods. *Statistics and Computing 10*, 25–37.

Jordan, M. I., Z. Ghahramani, T. S. Jaakkola, and L. K. Saul (1998). An introduction to variational methods for graphical models. In M. I. Jordan (Ed.), *Learning in Graphical Models*, pp. 105–162. Kluwer.

Jordan, M. I. and R. A. Jacobs (1994). Hierarchical mixtures of experts and the EM algorithm. *Neural Computation 6*(2), 181–214.

Ueda, N. and Z. Ghahramani (2002). Bayesian model search for mixture models based on optimizing variational bounds. *Neural Networks 15*(10), 1223–1241.

Waterhouse, S. (1997). *Classification and Regression using Mixtures of Experts*. Ph. D. thesis, Department of Engineering, Cambridge University, U.K.

Waterhouse, S., D. MacKay, and T. Robinson (1996). Bayesian methods for mixtures of experts. In M. C. M. D. S. Touretzky and M. E. Hasselmo (Eds.), *Advances in Neural Information Processing Systems*, pp. 351–357. MIT Press.